\pdfoutput=1
\documentclass[11pt]{article}
\usepackage{acl}
\usepackage{times}
\usepackage{latexsym}
\usepackage[T1]{fontenc}
\usepackage[utf8]{inputenc}
\usepackage{microtype}
\usepackage{graphicx}
\usepackage[normalem]{ulem}
\usepackage{adjustbox}
\usepackage{fancyhdr}

\author{Priyanka Sukumaran$^{1}$ \bf Conor Houghton$^{2,*}$ Nina Kazanina$^{1,3,*}$ \\
  $^{1}$School of Psychological Sciences, University of Bristol, UK \\
  $^{2}$Department of Computer Science, University of Bristol, UK\\
  $^{3}$Institute of Cognitive Neuroscience, National Research University Higher School of Economics, Moscow, Russia\\
  $^{*}$ Equal contribution\\
  \texttt{\{p.sukumaran, conor.houghton, nina.kazanina\}@bristol.ac.uk}}

\title{Do LSTMs See Gender? \\ Probing the Ability of LSTMs to Learn Abstract Syntactic Rules}
\begin{document}
\pagestyle{fancy}
\fancyhf{} 
\fancyhead[R]{Accepted at EMNLP 2022 Workshop BlackboxNLP}
\maketitle
\begin{abstract}
LSTMs trained on next word prediction can accurately perform linguistic tasks that require tracking long-distance syntactic dependencies. Notably, model accuracy approaches human performance on number agreement tasks \cite{Gulordava2018ColorlessHierarchically}. However, we do not have a mechanistic understanding of how LSTMs perform such linguistic tasks. Do LSTMs learn abstract grammatical rules, or do they rely on simple heuristics? Here, we test gender agreement in French which requires tracking both hierarchical syntactic structures and the inherent gender of lexical units. Our model is able to reliably predict long-distance gender agreement in two subject-predicate contexts: noun-adjective and noun-passive-verb agreement. The model showed more inaccuracies on plural noun phrases with gender attractors compared to singular cases, suggesting a reliance on clues from gendered articles for agreement. Overall, our study highlights key ways in which LSTMs deviate from human behaviour and questions whether LSTMs genuinely learn abstract syntactic rules and categories. We propose using gender agreement as a useful probe to investigate the underlying mechanisms, internal representations, and linguistic capabilities of LSTM language models.
\end{abstract}

\section{Introduction}
Recurrent neural networks such as Long Short-Term Memory networks (LSTMs) have had remarkable success in linguistic tasks requiring grammatical competence. However, there is a lack of mechanistic understanding of LSTMs' linguistic success, which may mimic or inform human language processing. Long-distance number agreement tasks have been used in various languages to test if LSTMs process hierarchical syntactic structures as humans do \cite{Linzen2016AssessingDependencies,Gulordava2018ColorlessHierarchically,Giulianelli2018UnderInformation}; we extend this finding to gender agreement in French. Gender differs from number in that it is an inherent property of a word, whereas the number of a word is chosen based on a speaker's message. Gender agreement takes place in multiple contexts, including subject-predicate agreement, both for singular and plural nouns (Table \ref{tab:testsets}). We aim to investigate how gender agreement is represented and generalised across contexts by language models. 
Here, we first focus on noun-adjective (NA) and noun-passive-verb agreement (NP) with varying number of distractor words. \textit{`la \textbf{robe} que j'aime est très \underline{bleue}/bleu'} (the \textbf{dress} that I like is very \underline{blue}) is an example where the feminine noun \textit{`robe'} agrees with the feminine adjective \textit{`bleue'} separated by five gender neutral distractors, see Table \ref{tab:testsets_dist} for more examples. Secondly, we test performance on phrases which include attractor nouns of opposing gender and number to the main noun. For all test cases, we compare performance on singular noun phrases which have informative gendered articles \textit{`le/la'}, to plural noun phrases for which the masculine and feminine articles are the same \textit{`les'}; thus forcing the LSTM to rely solely on the gender of the plural noun itself.

We show that LSTMs can perform robust long-distance gender agreement, but suffer more variation in performance on shorter phrases. We also find that models may be relying on other gender clues such as articles for agreement. Both findings deviate from how humans are believed to process agreement, and also lead us to question, as in \cite{mitchell2019lstms}, how abstract the syntactic rules learnt by the LSTM really are.
\section{Language Model}\label{sec:2}
The LSTM language model\footnote{The LSTM was implemented in Python 3.7 with Pytorch 1.2.0 and CUDA 10.2.} from \citet{Gulordava2018ColorlessHierarchically} was adopted and retrained on the French corpus from \citet{Mueller2020Cross-LinguisticModels}. The LSTM had two layers of 650 units, and was trained with a batch size of 128, dropout of 0.2, and initial learning rate of 20. The French corpus had 80 million tokens for training and 10 million tokens each for validation and testing. We cleaned the vocabulary of 50,000 most common tokens by removing capitalisation, punctuation and repeated tokens due to errors in accents, resulting in 42,000 tokens\footnote{Our data-sets and code: \url{https://github.com/prisukumaran23/lstm_fr/tree/main}}. The remaining tokens in the corpus were tagged as unknown with <unk>. The LSTM was trained on next-word prediction. In our grammatical tests, we counted success when the LSTM assigned a higher probability to the target word of the correct gender or number, rather than the ungrammatical alternative (Table \ref{tab:testsets}).

\section{Results}
Validation perplexity averaged over five best model initialisations was $43.57\pm0.18$. To verify our model's robustness, we first tested subject-verb number agreement in French \cite{Mueller2020Cross-LinguisticModels}. Our model had an overall accuracy of $95\%\pm0.08$, outperforming the model reported in \citet{Mueller2020Cross-LinguisticModels}, $83\%\pm0.18$, likely because we used a version of the corpus that was improved by removing duplicates. Our model performance follows the patterns in \cite{Mueller2020Cross-LinguisticModels}: high performance on simple number agreement without distractors, $100\%\pm0.03$, e.g. \textit{`les \textbf{pilotes} \underline{retournent}/retourne'}, and lowest performance on the across object relative clause condition, $71\%\pm0.11$, e.g. \textit{`les \textbf{auteurs} que les gardes aiment \underline{retournent}/retourne'}. 

\begin{figure}
  \centering
  \includegraphics[width=7.7cm]{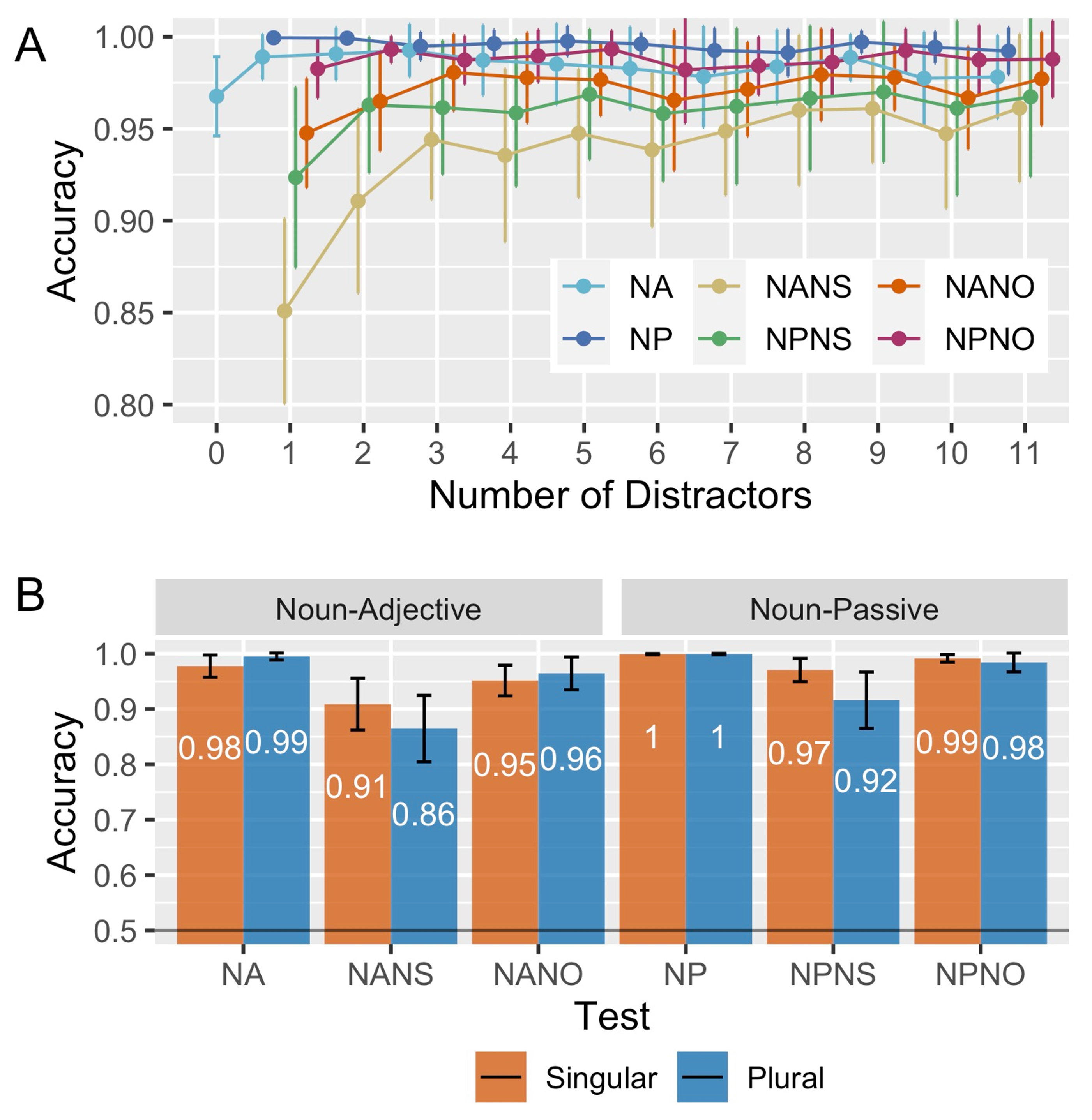}
  \caption{Model accuracy for each condition averaged across five best initialisations of the LSTM model. (A) Gender agreement performance on phrases with a varying number of gender neutral distractor words. (B) Average accuracy for each condition broken down for singular and plural phrases with 0-2 distractor words. Chance performance is marked at 0.5.}
\label{fig:results}
\end{figure}

Next, we tested gender agreement in phrases with an increasing number of distractor words between the noun and target word, which did not degrade agreement accuracy. Model performance had more variation on shorter phrases with 0-2 distractor words (Figure \ref{fig:results}A). Average accuracy on shorter phrases was very high for simple gender agreement without attractors, NA: $98\%$ for singular and $99\%$ for plural phrases, and NP: $100\%$ on both singular and plural conditions (Figure \ref{fig:results}B).
On phrases with gender attractors of the same number (NANS and NPNS), performance was $91\%$ and $97\%$ on singular phrases but drops to $86\%$ and $92\%$ in plural noun phrases which may be due to the absence of gendered articles.
Finally, accuracy was above $95\%$ and similar between singular/plural phrases phrases in the NANO and NPNO conditions, and
higher than in NANS and NPNS, possibly due to a lack of gender interference from articles.

\begin{table*}
\renewcommand{\arraystretch}{1.2}
\centering
\begin{adjustbox}{width=1\textwidth}
\small
\begin{tabular}{lll}
\hline \textbf{Test} & \textbf{Singular} & \textbf{Plural} \\ \hline
\textbf{\textit{Noun-Adjective Agreement:}}\\
\hline
\textit{NA}  & la \textbf{robe} est \underline{bleue}/bleu & les \textbf{robes} sont \underline{bleues}/bleus\\\vspace{0.2cm}
\textit{\small{(No Attractor)}} & the \textbf{dress}\textsubscript{f.s} is \uline{blue}\textsubscript{f.s}/blue\textsubscript{m.s} & the \textbf{dresses}\textsubscript{f.p} are \underline{blue}\textsubscript{f.p}/blue\textsubscript{m.p}\\

\textit{NANS} & la \textbf{robe} avec le sac est \underline{bleue}/bleu & les \textbf{robes} avec les sacs sont \underline{bleues}/bleus\\\vspace{0.2cm}
\textit{(Noun-Attractor Same number)} & the \textbf{dress}\textsubscript{f.s} with the bag\textsubscript{m.s} is \uline{blue}\textsubscript{f.s}/blue\textsubscript{m.s} & the \textbf{dresses}\textsubscript{f.p} with the bags\textsubscript{m.p} are \underline{blue}\textsubscript{f.p}/blue\textsubscript{m.p}\\

\textit{NANO} & la \textbf{robe} avec les sacs est \underline{bleue}/bleu & les \textbf{robes} avec le sac sont \underline{bleues}/bleus\\\vspace{0.3cm}
\textit{(Noun-Attractor Opposite number)} & the \textbf{dress}\textsubscript{f.s} with the bags\textsubscript{m.p} is \uline{blue}\textsubscript{f.s}/blue\textsubscript{m.s} & the \textbf{dresses}\textsubscript{f.p} with the bag\textsubscript{m.s} are \underline{blue}\textsubscript{f.p}/blue\textsubscript{m.p}\\
\textbf{\textit{Noun-Passive-Verb Agreement:}}\\
\hline
\textit{NP} & la \textbf{robe} est \underline{tombée}/tombé & les \textbf{robes} sont \underline{tombées}/tombés\\\vspace{0.2cm}
\textit{(No Attractor)} & the \textbf{dress}\textsubscript{f.s} \underline{fell}\textsubscript{f.s}/fell\textsubscript{m.s} & the \textbf{dresses}\textsubscript{f.p} \underline{fell}\textsubscript{f.p}/fell\textsubscript{m.p}\\

\textit{NPNS} & la \textbf{robe} avec le sac est \underline{tombée}/tombé & les \textbf{robes} avec les sacs sont \underline{tombées}/tombés\\\vspace{0.2cm}
\textit{(Noun-Attractor Same number)} & the \textbf{dress}\textsubscript{f.s} with the bag\textsubscript{m.s} \underline{fell}\textsubscript{f.s}/fell\textsubscript{m.s} & the \textbf{dresses}\textsubscript{f.p} with the bags\textsubscript{m.p} \underline{fell}\textsubscript{f.p}/fell\textsubscript{m.p}\\

\textit{NPNO} & la \textbf{robe} avec les sacs est \underline{tombée}/tombé & les \textbf{robes} avec le sac sont \underline{tombées}/tombés\\\vspace{0.2cm}
\textit{(Noun-Attractor Opposite number)} & the \textbf{dress}\textsubscript{f.s} with the bags\textsubscript{m.p} \underline{fell}\textsubscript{f.s}/fell\textsubscript{m.s} & the \textbf{dresses}\textsubscript{f.p} with the bag\textsubscript{m.s} \underline{fell}\textsubscript{f.p}/fell\textsubscript{m.p}\\
\hline
\end{tabular}
\end{adjustbox}
\caption{Test phrases for noun-adjective and noun-passive-verb gender agreement were constructed by systematically varying 40 nouns, 15 adjectives/passive-verbs, and 30 distractor phrases with up to 11 gender neutral words. We tested both singular and plural noun phrases for each condition. NA and NP conditions contain no gender-attractors, and we use the feminine noun, \textit{`robe'}(f.s) / \textit{`robes'}(f.p) as the main noun. The masculine noun, \textit{`sac'}(m.s) / \textit{`sacs'}(m.p) was used as the gender-attractor in the conditions NANS, NPNS, NANO and NPNO. The main noun determines the gender of the target adjective: \textit{`bleue'}(f.s) / \textit{`bleu'}(m.s) / \textit{`bleues'}(f.p) / \textit{`bleus'}(m.p) or passive-verb: \textit{`tombée'}(f.s) / \textit{`tombé'}(m.s) / \textit{`tombées'}(f.p) / \textit{`tombés'}(m.p), where f.s: feminine singular, m.s: masculine singular, f.p: feminine plural, m.p: masculine plural. The main noun is in bold and the correct form of the target adjective/passive-verb is underlined. Phrases shown here have one gender neutral word \textit{`est/sont'} between the nouns and agreement target, see Table \ref{tab:testsets_dist} for examples of long-distance agreement with varying number of distractors. }
\label{tab:testsets}
\end{table*}

\begin{table*}
\renewcommand{\arraystretch}{1.2}
\centering
\small
\begin{tabular}{lccl}
\hline \textbf{Test} & \textbf{Attractor} & \textbf{Distractors} & \textbf{Singular Sentence} \\ \hline
\textit{NA}
& - & 2 & la \textbf{robe} \textit{est très} \underline{bleue}/bleu\\\vspace{0.1cm} &&& the \textbf{dress}\textsubscript{f.s} is very \uline{blue}\textsubscript{f.s}/blue\textsubscript{m.s}\\
& - & 5 & la \textbf{robe} \textit{que j' aime beaucoup est} \underline{bleue}/bleu\\\vspace{0.1cm} &&& the \textbf{dress}\textsubscript{f.s} that I like is \uline{blue}\textsubscript{f.s}/blue\textsubscript{m.s}\\
\textit{NANS}
& 1 & 2 & la \textbf{robe} avec le sac \textit{est très} \underline{bleue}/bleu\\\vspace{0.1cm} &&& the \textbf{dress}\textsubscript{f.s} with the bag\textsubscript{m.s} is very \uline{blue}\textsubscript{f.s}/blue\textsubscript{m.s}\\
& 1 & 5 & la \textbf{robe} avec le sac \textit{que j' aime beaucoup est} \underline{bleue}/bleu\\\vspace{0.1cm} &&& the \textbf{dress}\textsubscript{f.s} with the bag\textsubscript{m.s} that I like is \uline{blue}\textsubscript{f.s}/blue\textsubscript{m.s}\\
\textit{NANO}
& 1 & 2 & la \textbf{robe} avec les sacs \textit{est très} \underline{bleue}/bleu\\\vspace{0.1cm} &&& the \textbf{dress}\textsubscript{f.s} with the bags\textsubscript{m.p}  is very \uline{blue}\textsubscript{f.s}/blue\textsubscript{m.s}\\
& 1 & 5 & la \textbf{robe} avec les sacs \textit{que j' aime beaucoup est} \underline{bleue}/bleu\\\vspace{0.1cm} &&& the \textbf{dress}\textsubscript{f.s} with the bags\textsubscript{m.p} that I like is \uline{blue}\textsubscript{f.s}/blue\textsubscript{m.s}\\
\hline
\end{tabular}
\caption{Example phrases for noun-adjective gender agreement with two and five gender neutral distractor words between the nouns and target adjective. The complete test-set contained phrases with up to 11 distractor words. Only singular noun-adjective agreement phrases are shown here, the equivalent plural phrases and noun-passive-verb agreement cases were also tested. The main noun is in bold, distractor words are italicised, and the correct form of the target adjective/passive-verb is underlined. Note: f.s: feminine singular, m.s: masculine singular, f.p: feminine plural, m.p: masculine plural.}
\label{tab:testsets_dist}
\end{table*}
\section{Discussion}
We find that LSTM language models produce robust gender agreement even with long distractor phrases and interfering attractors in French. Our LSTM struggled more on short phrases with attractors, deviating from human behaviour where longer phrases are found to incur more errors due to memory retrieval effects \cite{Alonso2021GenderComprehension}. We also find that the LSTM's performance on gender agreement differed for singular and plural gender associations; lesion studies could help identify specific gender units to confirm this \cite{Lakretz2019TheModels, Lakretz2021MechanismsHumans}. 

More work needs to be done to characterise the role of articles and other gender indicators in agreement. More broadly, we aim to explore contexts beyond noun-predicate agreement and whether learnt gender rules generalise across contexts and to novel nouns. Overall, gender agreement can be used to probe LSTMs' abilities to capture inherent grammatical categories and rules, in turn providing insight into the extent to which LSTMs inform us on the principles of language processing.

\newpage\thispagestyle{empty}

\bibliography{references}
\bibliographystyle{acl_natbib}
\end{document}